\pdfoutput=1

\documentclass[10pt,twocolumn,letterpaper]{article}

\usepackage{cvpr}

\definecolor{cvprblue}{rgb}{0.21,0.49,0.74}
\usepackage[pagebackref,breaklinks,colorlinks,allcolors=cvprblue]{hyperref}

\usepackage{booktabs}
\usepackage{multirow}
\usepackage{float}
\usepackage{caption}

\usepackage[table]{xcolor}
\definecolor{rowgray}{gray}{0.93}

\newcommand{\cmark}{\ensuremath{\checkmark}}
\newcommand{\xmark}{\ensuremath{\times}}

\def\confName{CVPR}
\def\confYear{2027}

\title{Tracktail: Multi-View 3D Point Tracking without Depth Input}

\author{Katie L. Rupp\\
Allen Institute, Seattle, WA\\
{\tt\small katie.rupp@alleninstitute.org}
\and
Jonathan Wong \\
Allen Institute, Seattle, WA\\
\and
Sharmishtaa Seshamani\\
Allen Institute, Seattle, WA\\
\and
Ana Mafalda Vicente\\
Allen Institute, Seattle, WA\\
\and
Lili Karashchuk\\
HHMI Janelia Research Campus, Ashburn, VA\\
{\tt\small karashchukl@janelia.hhmi.org}
}

\begin{document}

\twocolumn[{%
    \maketitle
    \begin{center}
        \hypertarget{fig:rat7m-top}{}%
        \includegraphics[width=0.9\linewidth]{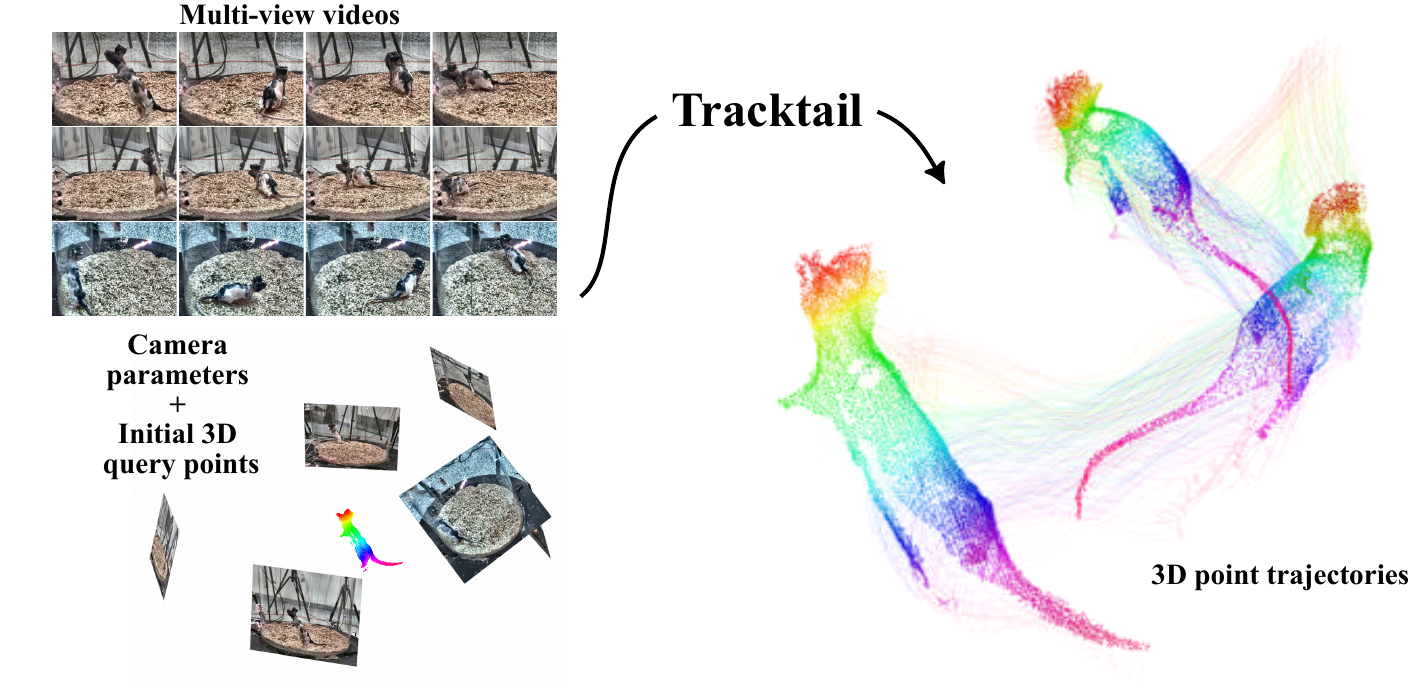}
        \captionof{figure}{We introduce Tracktail, a multi-view point tracker that tracks points in 3D without taking any depth input.}
        \label{fig:rat7m}
    \end{center}
}]

\begin{abstract}

We introduce Tracktail, a novel architecture designed to efficiently track query points in 3D from multiple calibrated views without any depth input. Our architecture separates the task as a per-camera video encoding of features and a per-query decoder. Notably, camera cross-attention is only performed across queries in the decoder, allowing for efficient inference (7 FPS and less than 4 GB memory for 10 views and 100 query points). We further introduce a per-camera scene scale computed from camera geometry, which allows us to normalize inputs and predictions appropriately across a vast variety of dataset scales. We train on a synthetic Multi-View Kubric dataset as well as a novel Multi-View Point Odyssey benchmark dataset, which consists of synthetic renders of humans and animals across 499 frames. Tracktail achieves state-of-the-art results on the Panoptic Studio, DexYCB, and Multi-View Kubric benchmarks, outperforming methods that rely on estimated depth. Tracktail also demonstrates a strong ability to maintain accurate tracking over longer ranges compared to existing methods. Our code, data, and weights are publicly available: \hyperlink{https://ai-hhmi.github.io/tracktail/}{https://ai-hhmi.github.io/tracktail/}

\end{abstract}
    
\section{Introduction}
\label{sec:intro}

Despite recent developments in 2D point tracking, long-term tracking of arbitrary points in 3D remains a fundamental problem in computer vision. Multi-View 3D point tracking has many applications such as dynamic scene reconstruction, robotics, and pose estimation. 

Current 3D tracking approaches typically rely on having access to depth information\cite{rajic2025mvtracker}. While methods like VGGT \cite{wang2025vggtvisualgeometrygrounded} or DUSt3R \cite{wang2024dust3rgeometric3dvision} can be used to acquire depth estimates, they are computationally expensive and often quite noisy, which can greatly hinder tracking performance. Existing methods are also unable to robustly predict on datasets with different units and scene scales, requiring arbitrary scaling per-dataset \cite{rajic2025mvtracker}. 

To overcome these challenges, we developed Tracktail, a multi-view 3D point tracking model that does not take any depth input and robustly handles various scene scales. Our model takes a multi-view video, corresponding camera parameters, and arbitrary query points as input (\hyperlink{fig:rat7m-top}{Fig.~\ref{fig:rat7m}}). A pretrained video encoder encodes the frames independently per view to obtain a latent scene representation, which is then passed through a decoder. The decoder leverages camera self-attention across views and temporal self-attention across frames, to produce predictions for 2D and 3D positions, visibility, depth, and confidences.  

\noindent Our key contributions are as follows: 
\begin{itemize}
    \item We introduce Tracktail, a depth-free multi-view 3D point tracker with an encoder-decoder architecture 
    \item We robustly handle the scene normalization problem by introducing a scene scale parameter computed from camera parameters and query points
    \item We establish a new benchmark dataset for 3D tracking, Multi-View Point Odyssey, to assess long-term tracking
    \item We demonstrate that our model achieves state-of-the-art results on existing benchmarks in both positional and occlusion accuracy. 
    \item We show that inference with our model is both faster and more memory efficient than existing 3D point tracking methods and scales better with the number of views 
\end{itemize}

\section{Related Work}
\label{sec:related_work}

\paragraph{Datasets.}
Many point tracking benchmarks have relatively short test sequences, which limits the extent to which long-term tracking can be assessed. For example, DexYCB and Multi-View Kubric only have 24 frames. Panoptic Studio has slightly longer sequences at 150 frames. The Point Odyssey benchmark \cite{zheng2023point} was created specifically to provide a long range tracking training dataset and benchmark for 2D point tracking, with the test videos averaging 2000 frames. More recently, Kubric-1024 was established as another long-range training dataset \cite{jung2026tapnextwhatstrackingpoint}. While Point Odyssey and Kubric-1024 are great for assessing long-range 2D point tracking, there is not yet an established benchmark to assess long-range multi-view point tracking.
 
We extend the Point Odyssey generation pipeline \cite{zheng2023point} to introduce a new benchmark dataset for 3D tracking, Multi-View Point Odyssey, consisting of human and animal datasets. The dataset consists of synthetic scenes with naturalistic motion sourced from motion capture. The test sequences are 499 frames, making it a useful benchmark for advancing long-term tracking. We trained our model on this new benchmark in addition to Multi-View Kubric, enabling our model to surpass state-of-the-art performance on real-world benchmarks.

\paragraph{2D Point Tracking.}
2D point tracking methods have greatly advanced over the past several years \cite{zheng2023point, karaev23cotracker, wang2023tracking, cho2024localallpaircorrespondencepoint, karaev24cotracker3, zhang2026d4rt}. One of the earliest 2D tracking methods, based on persistent independent particles (PIPs), was limited to an 8-frame window due to the MLP-mixer component of the architecture \cite{harley2022particlevideorevisitedtracking}. PIPs++ removes this 8-frame constraint by replacing the MLP-mixer with a CNN and adds a template-update strategy to account for changing appearance in objects over time, enabling longer range tracking \cite{zheng2023point}. CoTracker introduces an efficient transformer architecture and tracks points jointly to leverage spatial and temporal dependencies \cite{karaev23cotracker}. CoTracker3 simplifies CoTracker, improving efficiency and reducing the amount of data required to train the model \cite{karaev24cotracker3}. 
LocoTrack introduces a local 4D correlation to improve point correspondence, addressing point drift and ambiguities that are commonly encountered when tracking regions of uniform color or texture \cite{cho2024localallpaircorrespondencepoint}. 

\paragraph{3D Point Tracking.}
More recently, developments have been made in 3D point tracking for both monocular and multi-view approaches \cite{SpatialTracker, tapip3d, xiao2025spatialtrackerv2, luiten2023dynamic, rajic2025mvtracker, fradlin2026arrow}. SpatialTracker extends CoTracker by using monocular depth estimators to perform tracking in 3D through a triplane representation \cite{SpatialTracker}. SpatialTrackerV2 \cite{xiao2025spatialtrackerv2} and TAPIP3D \cite{tapip3d} also take monocular approaches, the latter utilizing spatio-temporal feature clouds. Dynamic 3DGS leverages 3D gaussian splatting to reconstruct scenes and track points in 3D \cite{luiten2023dynamic}, but requires optimizing over a large number of cameras (e.g. 27). MVTracker leverages depth and a 3D feature point cloud for multi-view 3D point tracking in a setup with a practical number of cameras (e.g. 4) \cite{rajic2025mvtracker}, but requires tuning arbitrary scene scaling factors empirically per dataset. D4RT uses an encoder-decoder architecture to encode a global scene representation and decode the 3D position of a 2D coordinate at a desired target time and camera \cite{zhang2026d4rt}. Most recently, ARROW takes inspiration from D4RT's query formulation to introduce a multi-view point tracker that incorporates global context \cite{fradlin2026arrow}. This is done through a global attention layer where every query attends to the full encoded context, making it computationally expensive \cite{fradlin2026arrow}. 

We extend the D4RT architecture to produce an efficient multi-view 3D point tracker that does not rely on depth information. Our method also directly addresses the scene scale problem by internally applying scene normalization that allows our model to robustly predict 3D positions and occlusions on a variety of datasets regardless of scale.

\section{Methodology}
\label{sec:methodology}

\begin{figure*}[t]
    \centering
    \includegraphics[width=0.99\linewidth]{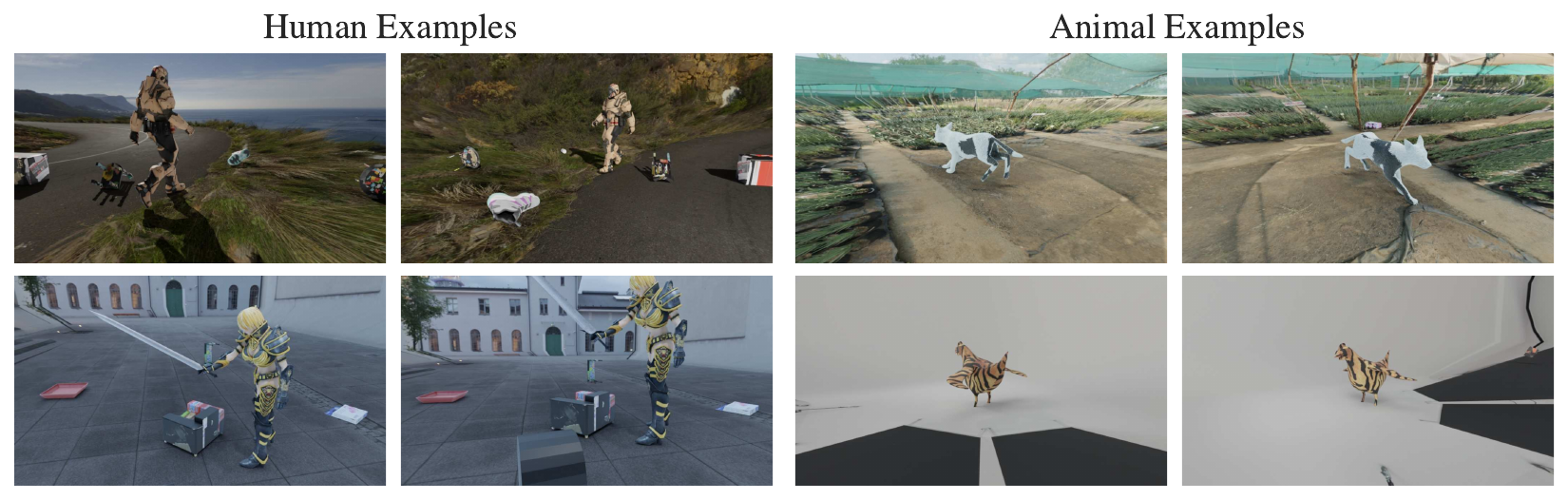}
    \caption{\textbf{Multi-View Point Odyssey datasets} - We introduce two Multi-View Point Odyssey datasets, one with human subjects and the other with animal subjects. We show two example frames from human (right) and animal (left) trials. Each trial has 9 static camera views and is generated by randomly sampling a subject, an HDRI environment, a motion trajectory, and 5 distractor objects. Random fur textures are sampled for the animal dataset as well.}
    \label{fig:po_examples}
\end{figure*}

\subsection{Multi-View Point Odyssey Dataset}

We extend the Point Odyssey generation pipeline \cite{zheng2023point} to render each scene (\cref{fig:po_examples}) from 9 static cameras placed around the subject, with ground-truth 3D trajectories and per-camera visibility for every surface point. The human dataset consists of Mixamo characters \cite{mixamo} that are rigged with AMASS motions \cite{mahmood2019amass}, and the animal dataset consists of DeformingThings4D animal models and animation sequences \cite{li20214dcomplete} with random fur textures from BlenderKit. Each scene has a random HDRI environment and distractor objects. Random forces are applied to these objects to push them around throughout the video. The human dataset has 113 training and 8 test sequences, and the animal dataset has 210 training and 10 test sequences. Each test sequence is 499 frames long and comes with 512 evaluation tracks stratified by speed. See Appendix \ref{sec:supp_po} for further details.

\begin{figure*}[t]
    \centering
    \includegraphics[width=0.95\linewidth]{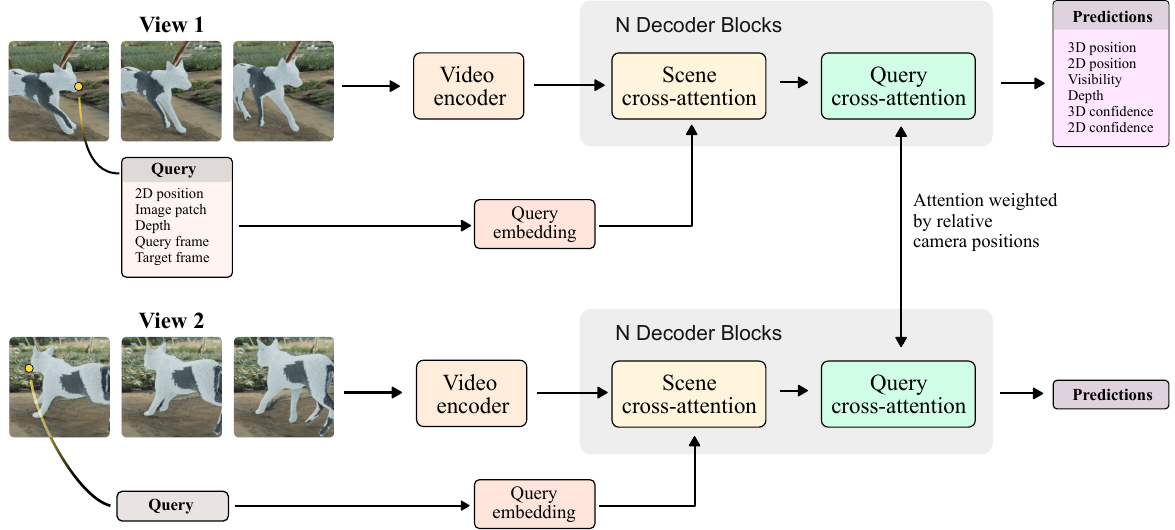}
    \caption{\textbf{Tracktail architecture} - Given frames from multi-view videos, the corresponding camera intrinsics and extrinsics, and 3D query points with their query times, Tracktail predicts the 2D and 3D trajectories and visibilities of the query points for all input frames. First, the frames are passed through a video encoder (V-JEPA 2.1) independently per view to obtain a scene representation. The 3D queries are passed through a query encoder to obtain a camera-specific query embedding for each view and target frame. These embeddings are processed by a series of decoder blocks with camera self-attention across views, temporal self-attention across frames, and cross-attention from each view's queries to that view's scene tokens. The decoder predicts per-camera 3D positions, 2D positions, depth, visibility, and fusion weights, which are combined across cameras into the final 3D trajectories.}
    \label{fig:architecture}
\end{figure*}

\subsection{Problem formulation}

We follow the problem formulation of MVTracker \cite{rajic2025mvtracker}. Given a set of synchronized RGB videos from static cameras with known parameters and a set of 3D query points with query times, the task is to predict the position of each query point at every target time. That is, we are given:
\begin{itemize}
  \item a set of video sequences $\mathbf{V}^c$ composed of images $\{\mathbf{I}^c_t \in \mathbb{R}^{H \times W \times 3} \; | \; t \in T \}$
  \item the query points $\mathbf{q}_n = ( x_n, y_n, z_n)$ and query frames $t_n$
  \item the calibration parameters $\mathbf{M}_c = (\mathbf{K}_c, \mathbf{E}_c, \mathbf{d}_c, \mathbf{o}_c)$ including intrinsics $\mathbf{K}_c \in \mathbb{R}^{3 \times 3}$, extrinsics $\mathbf{E}_c \in \mathbb{R}^{4 \times 4}$, distortions $\mathbf{d}_c \in \mathbb{R}^5$, and crop offsets $\mathbf{o}_c \in \mathbb{R}^{2}$
\end{itemize}
where $c$ is the camera number, $t$ is the frame number, and $n$ is the point number. No depth maps, depth estimates, or visibility information are given as input.

The task is to predict the 3D trajectory $\{\mathbf{p}^t_n = (x_n^t, y_n^t, z_n^t) \, | \, t \in T\}$ along with visibilities $v_n^t \in \{0, 1\}$, where $v_n^t = 1$ indicates that a point is visible in at least one camera view.

\subsection{Scene scale}

A key challenge in multi-view point tracking is normalizing across cameras with different effective scales and across datasets with different units. We address this with a per-camera scene scale $s_c$, which is computed analytically from the camera parameters and the query points (it is not learned).

For each camera matrix $\mathbf{E}_c$ and query point $\mathbf{q}_n$, let $\mathbf{E}_c \tilde{\mathbf{q}}_n$ be the point in camera coordinates with components $(X, Y, Z)$. The Jacobian of the projection (extrinsics followed by pinhole projection) is:

\begin{align}
    \mathbf{J}_c(\mathbf{q}_n) = \begin{pmatrix} \frac{f_x}{Z} & 0 & -\frac{f_x X}{Z^2} \\[2pt] 0 & \frac{f_y}{Z} & -\frac{f_y Y}{Z^2} \end{pmatrix} \mathbf{R}_c  
\end{align}
where $\mathbf{R}_c$ is the rotation part of $\mathbf{E}_c$, so that $\mathbf{J}_c \in \mathbb{R}^{2 \times 3}$.

The largest singular value $\sigma_{\max}(\mathbf{J}_c(\mathbf{q}_n))$ is the maximum pixel displacement per unit of 3D motion at $\mathbf{q}_n$. The per-camera scene scale is its inverse, aggregated over the query points

\begin{align}
s_c = 1/\underset{n \in \mathcal{V}_c}{\text{median}} \; \sigma_{\max}(\mathbf{J}_c(\mathbf{q}_n))
\end{align}

where $\mathcal{V}_c$ is the set of query points that lie in front of camera $c$ and project inside its image. For cameras with no such query points, $s_c$ is set to the median over the remaining cameras.

Thus, $s_c$ is the 3D displacement (in scene units) that moves a point by about one pixel in camera $c$. We use $s_c$ to express all geometric quantities seen and produced by the network in pixel-like units: the query depth input is divided by $s_c$, the predicted 3D motion is expressed in units of $s_c$, and the predicted depth in units of $s_c f_c$, where $f_c$ is the focal length. The 3D losses are normalized the same way. This allows joint training and inference across camera rigs and datasets with arbitrary units without any manual rescaling.

\subsection{Model architecture}

Our architecture extends D4RT \cite{zhang2026d4rt} to multi-view 3D point tracking (\cref{fig:architecture}). A video encoder $\mathcal{E}$ encodes each video $\mathbf{V}^c$ independently into a scene representation $\mathbf{F}^c = \mathcal{E}(\mathbf{V}^c)$. We use the V-JEPA 2.1 ViT-L/16 encoder \cite{murlabadia2026vjepa2_1} on 24-frame clips at $256 \times 256$ pixels, with hierarchical features projected to 512 dimensions.
A query encoder $\mathcal{Q}$ produces an embedding $\mathbf{z}_{n,\tau}^c = \mathcal{Q}(\mathbf{q}_n, t_n, \tau, \mathbf{M}_c, \mathbf{I}^c_{t_n})$ for each camera $c$, query point $n$, and target time $\tau$, from the query's projection into the camera, an image patch around it, its depth relative to $s_c$, the camera intrinsics, the query and target times, and an optional per-camera occlusion state.

A decoder $\mathcal{D}$ with 6 transformer blocks then processes the embeddings. Each block applies camera self-attention across views, temporal self-attention across target frames, and cross-attention from each view's queries to that view's scene tokens. Camera self-attention uses camera ray encodings $\mathbf{r}^c_n$ as relative positional encodings, following PRoPE \cite{li2025camerasrelativepositionalencoding,zhang2026unifiedcamerapositionalencoding}. Camera self-attention is the only place where information is exchanged between views, and different query points never attend to each other, so the model can track an arbitrary number of points.

\paragraph{Predictions and multi-view fusion.} For each camera, query point, and target time, the decoder predicts a 3D position, a 2D position, a depth, a visibility logit $\ell^c_{n,\tau}$, and a 3D fusion logit. Positions are predicted as per-axis classifications over fixed bins decoded by a soft-argmax, similar to TAPNext \cite{zholus2025tapnext}, with the 3D position predicted as a motion relative to the query position in units of $s_c$. The final 3D prediction is a weighted average of the per-camera predictions, $\mathbf{p}_{n,\tau} = \sum_c w^c_{n,\tau} \mathbf{p}_{n,\tau}^c$, where $w^c_{n,\tau}$ is the softmax over cameras of the fusion logits. The overall visibility is the noisy-OR of the per-camera visibility probabilities $v^c_{n,\tau} = \sigma(\ell^c_{n,\tau})$, $v_{n,\tau} = 1 - \prod_{c} (1 - v^c_{n,\tau})$, which matches the definition that a point is visible if it is visible in at least one camera. The 2D and depth predictions are used for auxiliary supervision.
The same network also accepts 2D pixel queries in a single view, which we use as an auxiliary training task. See Appendix \ref{sec:supp_architecture} for further details.

\subsection{Inference on long videos}

At inference, each view is cropped to a box around the projections of the query points, with a 20 pixel margin and a minimum size of 256 pixels. The crop is resized so that its longer side is 256 pixels and zero-padded to $256 \times 256$, and the intrinsics and crop offsets $\mathbf{o}_c$ are updated accordingly.

Longer videos are processed in sliding windows of 24 frames that overlap by 8 frames. Each point is first queried in the window that contains its query frame. In each following window it is re-queried at the last overlapping frame, using the previous window's 3D prediction as the query position. Each window is re-cropped around the current query positions, expanded by the points' recent velocity so that fast-moving subjects stay in view. Points predicted to be invisible are excluded from the crop. For re-queried points, the occlusion state input is set from the previous window's per-camera visibility predictions, thresholded at 0.1. 

\section{Experiments}
\label{sec:experiments}

\subsection{Datasets}
\label{sec:datasets}

\paragraph{Training data.} We train only on synthetic data from three multi-view datasets with static, calibrated cameras:
(i) Multi-View Kubric (our renders) \cite{greff2021kubric,rajic2025mvtracker}: 4,989 scenes with 10 cameras, 24 frames each, at $512 \times 512$
(ii) Multi-View Point Odyssey Human: 113 sequences, and
(iii) Multi-View Point Odyssey Animal: 210 sequences.
We sample 24-frame training clips with a temporal stride of 1, 2, or 4. Datasets are balanced by the number of available clips.

\paragraph{Evaluation data.} We evaluate on the three multi-view benchmarks of MVTracker \cite{rajic2025mvtracker}, using its four input cameras and its fixed track sets:
Multi-View Kubric (30 test sequences of 24 frames, 512 tracks each), DexYCB \cite{chao2021dexycb} (10 sequences of 24 frames, 384 tracks each), and Panoptic Studio \cite{joo2015panoptic} (6 sequences of 150 frames, 384 tracks each).
Each track is queried at the frame given by the benchmark.
To assess long-range tracking, we introduce Multi-View Point Odyssey. It consists of the test splits of human (8 sequences) and animal (10 sequences) datasets, each 499 frames long, with 512 tracks per sequence stratified by speed. Each track is queried at the first frame where it is visible in the cameras used. We evaluate with a subset of 4 cameras (views 2, 4, 5, and 7) and with all 9 cameras.
None of these benchmarks provide depth or visibility as input to Tracktail.

\subsection{Evaluation Metrics}
\label{sec:evaluation}

We follow the evaluation protocol of MVTracker \cite{rajic2025mvtracker} and compute all metrics in 3D world coordinates, on the frames after each track's query frame. A point is visible if it is visible in at least one input camera. We report the average Jaccard (AJ), which jointly measures position and visibility. We also report the position accuracy $\delta_\text{avg}$, averaged over a set of distance thresholds per dataset, the occlusion accuracy (OA), and the median trajectory error (MTE). On the long Point Odyssey sequences, we also report the survival rate of tracks over time. Each method uses its own visibility threshold (0.1 for Tracktail, see Appendix \ref{sec:supp_eval} for details).

\subsection{Training}
\label{sec:training}

\paragraph{Losses.} The training loss combines Huber losses on the fused and per-camera 3D predictions, normalized by the scene scale, with auxiliary Huber losses on the per-camera 2D, depth, and reprojected 3D predictions, cross-entropy losses on the 3D, 2D, and depth bins, binary cross-entropy losses on the per-camera and overall visibility, and a temporal smoothness penalty. The full loss and its weights are given in Appendix \ref{sec:supp_losses}.

\paragraph{Optimization.} We use Muon \cite{jordan2024muon} for the weight matrices of the transformer blocks and schedule-free AdamW \cite{defazio2024roadlessscheduled} for all other parameters. Both use schedule-free averaging, weight decay 0.002, a short linear warmup, and no learning rate decay. We train in two stages with PyTorch Lightning Fabric and distributed data parallelism in fp32, with one clip per GPU per iteration. We count iterations as the total number of clips processed across all GPUs:
\begin{enumerate}
    \item \textbf{Pretraining.} The V-JEPA encoder is frozen and we train the query encoder and decoder for 800k iterations on 4 NVIDIA H100 GPUs, which takes 49 hours. The learning rate is $10^{-3}$.
    \item \textbf{Finetuning.} We unfreeze the encoder and train end-to-end for 2.3M iterations on 8 H100 GPUs, which takes 8 days. The learning rate is $2.5 \times 10^{-4}$, with a $0.25\times$ learning rate for the encoder, and Multi-View Kubric clips are upweighted $3\times$.
\end{enumerate}
In total, training takes about 1,700 H100 GPU-hours. The model has 351M parameters: 308M in the video encoder and projection, 6M in the query encoder, and 36M in the decoder. We also leverage sampling and augmentations during training to improve model robustness (see Appendix \ref{sec:supp_training}).

\subsection{Comparison to Existing Methods}
\label{sec:comparison}

We compare Tracktail against monocular and multi-view 3D trackers on Panoptic Studio, DexYCB, and Multi-View Kubric in \cref{tab:main_results}. All methods use the same cameras, tracks, and query frames. We ran the following methods with our evaluation code:
\begin{itemize}
    \item \textbf{MVTracker} \cite{rajic2025mvtracker} with its released weights, using either ground-truth depth or depth estimated per frame with DUSt3R, with the camera parameters fixed to the ground truth. For DexYCB, the ground-truth depth is the Kinect sensor depth.
    \item \textbf{ARROW} \cite{fradlin2026arrow}. ARROW predicts depth up to scale, so we fit its scale to the ground truth once per 24-frame window and provide the ground-truth pose of the first camera.
    \item \textbf{CoTracker3 baseline}, a depth-free baseline that we construct. We project each 3D query into every view, track it in 2D with CoTracker3 \cite{karaev24cotracker3} (offline model), and triangulate the per-view tracks with linear triangulation using the views where it is predicted visible. If fewer than two views see the point, no prediction is made. See Supplementary Tab. \ref{tab:cotracker_triangulation} for an analysis of these failures.
\end{itemize}
We also compare against the above methods on Multi-View Point Odyssey (\cref{tab:po_results}).

\begin{table*}[t!]
\centering
\caption{\textbf{Quantitative comparison of 3D point tracking performance across multiple benchmarks} - Results on Panoptic Studio, DexYCB, and Multi-View Kubric with 4 input cameras. We ran the methods using the same tracks (384 on Panoptic Studio and DexYCB, 512 on Multi-View Kubric), query frames, and cameras. Tracktail, which uses no depth input, achieves the best results on Panoptic Studio and DexYCB, outperforming MVTracker on AJ, $\delta_\text{avg}$, and MTE even when MVTracker is given ground-truth or sensor depth. On Multi-View Kubric, it is second only to MVTracker with ground-truth depth, and it outperforms MVTracker with DUSt3R-based depth on all datasets. The MTE is reported in centimeters (see Appendix \ref{sec:supp_eval} for units on Multi-View Kubric). $^\dagger$Kinect sensor depth.}
\label{tab:main_results}
\resizebox{\textwidth}{!}{%
\begin{tabular}{ll cccc cccc cccc}
\toprule
\multirow{2}{*}{Method} & \multirow{2}{*}{Train} 
  & \multicolumn{4}{c}{Panoptic Studio~\cite{joo2015panoptic}}
  & \multicolumn{4}{c}{DexYCB~\cite{chao2021dexycb}}
  & \multicolumn{4}{c}{Multi-View Kubric~\cite{rajic2025mvtracker}} \\
\cmidrule(lr){3-7}\cmidrule(lr){7-10}\cmidrule(lr){11-14}
  & & AJ$\uparrow$ & $\delta_\text{avg}\uparrow$ & OA$\uparrow$ & MTE$\downarrow$
      & AJ$\uparrow$ & $\delta_\text{avg}\uparrow$ & OA$\uparrow$ & MTE$\downarrow$
      & AJ$\uparrow$ & $\delta_\text{avg}\uparrow$ & OA$\uparrow$ & MTE$\downarrow$ \\
\midrule

\rowcolor{rowgray}
MVTracker (DUSt3R depth)~\cite{rajic2025mvtracker}      & MV-Kub           &  58.8 & 70.0 & 91.2 & 10.2 & 73.4 & 82.3 & \underline{91.3} & 2.0 & 51.5 & 67.2 & 79.0 & 3.7  \\
MVTracker (GT Depth) ~\cite{rajic2025mvtracker}                      & MV-Kub           &  \underline{84.2} & \underline{93.6} & 91.4 & \underline{3.4} & \underline{83.0}$^\dagger$ & \underline{91.4}$^\dagger$ & \textbf{91.5}$^\dagger$ & \underline{0.8}$^\dagger$ & \textbf{80.8} & \textbf{89.6} & \textbf{93.5} & \textbf{0.7} \\
\midrule
\rowcolor{rowgray}
CoTracker3 Baseline               & Kub+15k     &   62.5 & 70.3 & 81.5 & 7.6 & 66.4 & 72.8 & 84.4 & 3.7 & 42.9 & 48.8 & 74.5 & 7.8 \\
ARROW~\cite{fradlin2026arrow} & See~\cite{fradlin2026arrow} &  69.1 & 77.6 & \underline{92.0} & 8.1 & 49.6 & 55.6 & 90.9 & 20.1 & 30.0 & 35.2 &
 88.7 & 9.2 \\
 \rowcolor{rowgray}
Tracktail (Ours)                & MV-Kub+MV-PO     &  \textbf{86.5} & \textbf{95.5} & \textbf{92.7} & \textbf{2.7} & \textbf{84.9} & \textbf{93.8} & \textbf{91.5} & \textbf{0.6} & \underline{77.4} & \underline{87.0} & \underline{92.3} & \underline{1.1} \\
\bottomrule
\end{tabular}%
}
\end{table*}

\begin{table*}[t!]
\centering
\caption{\textbf{Multi-View Point Odyssey Benchmark}. Results on the Multi-View Point Odyssey Human and Animal test sets (499 frames, 512 tracks per sequence) with a subset of 4 views (views 2, 4, 5, and 7) and with all 9 views. Tracktail outperforms the baseline methods for all evaluation metrics. MTE is reported in Point Odyssey scene units (see Appendix \ref{sec:supp_eval}). Note that Tracktail was trained on the corresponding training sets, while the other methods were not.}
\label{tab:po_results}
\resizebox{\textwidth}{!}{%
\begin{tabular}{l cccc cccc cccc cccc}
\toprule
\multirow{3}{*}{Method}
  & \multicolumn{8}{c}{4 Camera Views}
  & \multicolumn{8}{c}{9 Camera Views (all)} \\
\cmidrule(lr){2-9}\cmidrule(lr){10-17}
  & \multicolumn{4}{c}{Human}
  & \multicolumn{4}{c}{Animal}
  & \multicolumn{4}{c}{Human}
  & \multicolumn{4}{c}{Animal} \\
\cmidrule(lr){2-5}\cmidrule(lr){6-9}\cmidrule(lr){10-13}\cmidrule(lr){14-17}
  & AJ$\uparrow$ & $\delta_\text{avg}\uparrow$ & OA$\uparrow$ & MTE$\downarrow$
  & AJ$\uparrow$ & $\delta_\text{avg}\uparrow$ & OA$\uparrow$ & MTE$\downarrow$
  & AJ$\uparrow$ & $\delta_\text{avg}\uparrow$ & OA$\uparrow$ & MTE$\downarrow$
  & AJ$\uparrow$ & $\delta_\text{avg}\uparrow$ & OA$\uparrow$ & MTE$\downarrow$ \\
\midrule
\rowcolor{rowgray}
MVTracker~\cite{rajic2025mvtracker} (DUSt3R depth)
  & 8.2 & 11.1 & 64.4 & 16.2 & 8.7 & 13.8 & 53.2 & 19.7 & 12.6 & 16.0 & 66.3 & 14.5 & 12.4 & 18.7 & 68.4 & 13.2 \\
MVTracker~\cite{rajic2025mvtracker} (GT depth)
  & 17.0 & 23.0 & 66.9 & 8.2 & 14.3 & 21.0 & 60.0 & 9.1 & 22.6 & 29.1 & 69.8 & 7.1 & 21.6 & 30.2 & 65.8 & 5.8 \\
  \midrule
  \rowcolor{rowgray}
CoTracker3 Baseline~\cite{karaev24cotracker3}
  & 22.4 & 24.4 & 52.8 & \underline{4.5} & 25.4 & 28.7 & 52.0 & \underline{2.1} & 24.5 & 27.0 & 71.0 & 10.6 & 27.8 & 33.1 & 70.0 & 3.2 \\
  ARROW~\cite{fradlin2026arrow}
     & \underline{28.4} & \underline{34.1} & \underline{91.2} & 5.2 & \underline{38.6} & \underline{45.3} & \underline{93.5} & 3.2 & \underline{37.6} & \underline{42.2} & \underline{95.9} & \underline{4.2} & \underline{36.4} & \underline{43.6} & \underline{96.5} & \underline{3.1} \\
     \rowcolor{rowgray}
Tracktail (Ours)
  & \textbf{54.2} & \textbf{61.1} & \textbf{93.5} & \textbf{1.6} & \textbf{60.9} & \textbf{69.0} & \textbf{95.1} & \textbf{0.9} & \textbf{60.9} & \textbf{66.5} & \textbf{97.0} & \textbf{1.1} & \textbf{69.7} & \textbf{75.6} & \textbf{98.4} & \textbf{0.7} \\
\bottomrule
\end{tabular}%
}
\end{table*}

\paragraph*{Results on existing benchmarks.}

Table \ref{tab:main_results} shows the evaluation results on Panoptic Studio, DexYCB, and Multi-View Kubric with 4 input cameras. While our tracker tends to achieve only slightly better OA in comparison to existing methods, our model especially advances position-based metrics on these real-world benchmark datasets, setting a new state-of-the-art on Panoptic Studio and DexYCB. Notably, Tracktail achieves higher performance than MVTracker with GT depth on both Panoptic Studio and DexYCB despite not taking any depth input. It also performs considerably better than MVTracker with DUSt3R depth on all 3 existing benchmarks, emphasizing that noisy depth estimates can limit tracking performance. 

While MVTracker with GT depth achieves the highest performance on Multi-View Kubric, our model comes close while still performing considerably better than other existing methods. This makes sense, as MVTracker purely trains on Multi-View Kubric, while we train on a combination of Multi-View Kubric and Multi-View Point Odyssey. Training on the latter allows our model to generalize better to real world datasets, making a slight drop in performance on Multi-View Kubric a reasonable trade-off.  

\begin{figure*}[t]
    \centering
    \includegraphics[width=0.95\linewidth]{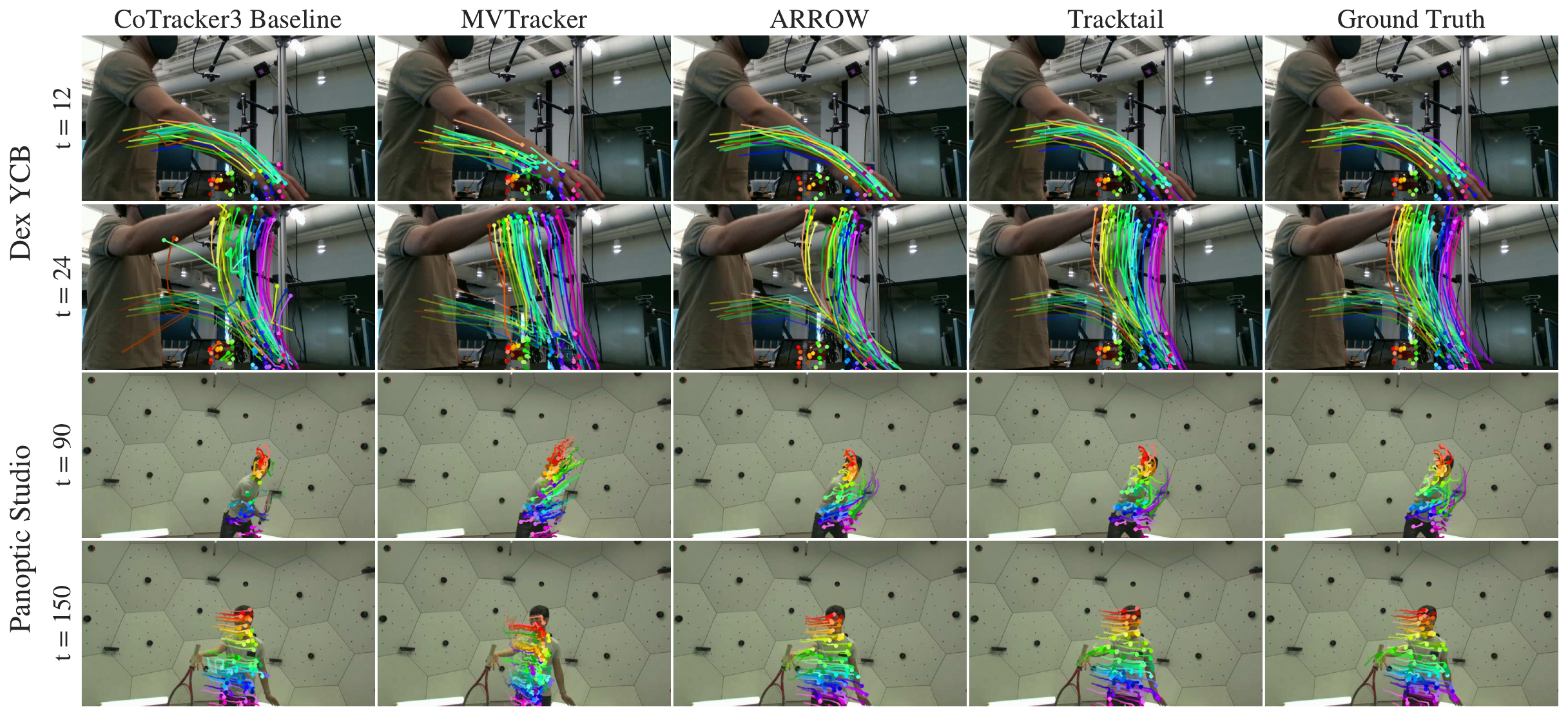}
    \caption{\textbf{Qualitative comparison} - Predicted trajectories of Tracktail, MVTracker, ARROW, and the CoTracker3 baseline compared to the ground truth on DexYCB and Panoptic Studio. Points are shown only when visible in the ground truth. Our method shows higher accuracy and smoother trajectories compared to baseline methods. All methods use the same 100 query points and the same 4 input views.}
    \label{fig:qualitative_comparison}
\end{figure*}

Figure \ref{fig:qualitative_comparison} shows a qualitative comparison between Tracktail and the baseline methods on DexYCB and Panoptic Studio that corresponds to the results in Table \ref{tab:main_results}. On DexYCB, the CoTracker3 baseline produces erratic trajectories, MVTracker's trajectories don't follow the proper direction of motion, and ARROW fails to track certain points. Our tracker produces point trajectories that match closest to the ground truth, visually showcasing Tracktail's strong performance on DexYCB. For Panoptic Studio, the CoTracker3 baseline doesn't capture the full range of motion, MVTracker produces very shaky trajectories, and ARROW shows some point drift in comparison to the ground truth. Tracktail also exhibits a little point drift, though this is expected since the videos are 150 frames.

\paragraph{Results on Multi-View Point Odyssey.}

Table \ref{tab:po_results} reports the performance of Tracktail and baseline methods on the Multi-View Point Odyssey test set. Tracktail outperforms all baseline methods for all metrics. This is somewhat expected, as we were the only one who trained on Multi-View Point Odyssey. ARROW struggled on existing benchmarks (Tab. \ref{tab:main_results}), but it generally outperforms the other baselines on this dataset, achieving an MTE of 3.1 on the animal set with all cameras (Tab.~\ref{tab:po_results}). This could be in part due to ARROW's training mixture, which includes the original single-view Point Odyssey dataset \cite{fradlin2026arrow}. MVTracker with GT depth consistently performs better than MVTracker with DUSt3R depth, again highlighting the importance of depth-free point tracking methods. In Appendix \ref{sec:supp_survival} and Supplementary Figures \ref{fig:po_survival_comparison} and \ref{fig:po_survival_tracktail}, we report survival rate on Multi-View Point Odyssey further quantify the survival of points over the 499-frame test videos. 

\paragraph{Comparative speed and memory usage.}
Supplementary Figure \ref{fig:speed_memory} compares the inference speed and memory usage of Tracktail with the baseline 3D tracking methods. For 4 views and 100 query points, Tracktail processes ~75$\times$ more frames per second compared to MVTracker with DUSt3R depth and ~3.5$\times$ more frames per second compared to ARROW (Fig \ref{fig:speed_memory}). Tracktail also maintains the lowest peak memory usage across all baseline methods, only slightly increasing with the number of camera views. In contrast, memory usage scales linearly with the number of views for MVTracker with DUST3R depth and ARROW. Figure \ref{fig:speed_memory_tracktail} shows that Tracktail's memory usage is primarily driven by the number of query points, while inference speed only slightly decreases. These results demonstrate that Tracktail is more efficient than existing multi-view 3D point tracking methods in terms of both speed and memory.

\subsection{Ablations}

We conducted ablation experiments to assess the effect of camera self-attention, number of input views, video encoder size, and latent dimension. Our final finetuned model was used to assess the number of input views. Our pretrained models, which were trained only for 800k iterations with a frozen encoder, were used for all other ablations.

\paragraph*{Effect of camera and temporal self-attention.}

Table \ref{tab:ablation_camattn} shows the effect of camera and temporal self-attention on Multi-View Kubric. Using camera self-attention improves performance across all metrics, suggesting that it is an essential part of our architecture. Including camera self-attention improves the MTE substantially from 5.1 to 2.8. AJ and $\delta_{avg}$ also show great improvements. These results emphasize the importance of the exchanging information across views in a 3D tracking setting. Temporal self-attention has a smaller effect: removing it lowers AJ and $\delta_{avg}$ by a little, while the MTE slightly improves. Removing it together with camera self-attention lowers AJ further. See Appendix \ref{sec:supp_eval} for details on how the evaluation was done.

\begin{table}[t]
\centering
\caption{\textbf{Ablation of temporal and camera self-attention} - Removing temporal self-attention slightly lowers AJ, $\delta_\text{avg}$, and OA on Multi-View Kubric, with a similar MTE. Removing camera self-attention causes a much larger drop on all metrics and removing both lowers AJ and $\delta_\text{avg}$ further. All models were trained with the pretraining stage only (800k iterations, frozen encoder).}
\label{tab:ablation_camattn}
\begin{tabular}{cc cccc}
\toprule
\multirow{2}{*}{\shortstack{Temporal\\Self-Attn.}} & \multirow{2}{*}{\shortstack{Camera\\Self-Attn.}}
  & \multicolumn{4}{c}{Multi-View Kubric~\cite{rajic2025mvtracker}} \\
\cmidrule(lr){3-6}
  & & AJ$\uparrow$ & $\delta_\text{avg}\uparrow$ & OA$\uparrow$ & MTE$\downarrow$ \\
\midrule
\cmark & \cmark & \textbf{60.2} & \textbf{70.4} & \textbf{89.4} & 2.8 \\
\rowcolor{rowgray}
\xmark & \cmark & 58.3 & 68.8 & 89.1 & \textbf{2.7} \\
\cmark & \xmark & 49.4 & 59.4 & 86.7 & 5.1 \\
\rowcolor{rowgray}
\xmark & \xmark & 46.3 & 56.4 & 86.8 & 5.2 \\
\bottomrule
\end{tabular}
\end{table}

\paragraph*{Effect of the number of input views.}

In Supplementary Figure \ref{fig:ncams}, we report the AJ on DexYCB for Tracktail and other 3D point trackers when given 1-8 camera views as input. Every method received the same combination of input views. Our model outperforms all methods under all view counts and generally improves with increasing view count. MVTracker with DUSt3R depth shows the most improvement as the number of views increases, but maintains a lower AJ compared to Tracktail. The CoTracker3 baseline doesn't show much improvement with increasing number of views, likely due to triangulation failures (Tab. \ref{tab:cotracker_triangulation}) and triangulation errors accumulating as views are added. ARROW shows the weakest performance on this dataset and doesn't have a steady increase in AJ beyond 3 views.

\paragraph*{Effect of the video encoder size.}
We assessed performance on Multi-View Kubric across the four V-JEPA 2.1 video encoder sizes, ranging from the base model with 86M parameters to the gigantic model with 2B parameters (Supplementary Table \ref{tab:ablation_encoder}). The largest increase in performance is seen when upgrading from the base model to the large model (300M parameters), as the MTE improves from 3.4 to 2.8. Performance further improves across all metrics for the giant model (1B parameters). AJ and $\delta_{avg}$ continue to increase with the gigantic model, suggesting that the giant and gigantic models are well-suited for applications where accuracy is the priority. Based on these results, we use the large model throughout.  

\paragraph*{Effect of the latent dimension.}
Supplementary Table \ref{tab:ablation_latent} shows the effect of the latent dimension on Tracktail performance on Multi-View Kubric. Increasing the latent dimension from 256 to 512 improves all evaluation metrics, with the MTE improving from 3.2 to 2.8. Further increasing the latent dimension to 1024 does not result in any improvement. These results show that a latent dimension of 512 offers the best trade-off in terms of accuracy and efficiency.

\section{Discussion and Future Work}
\label{sec:discussion}

One limitation of our method is that it does not account for camera movement, making accurate tracking on moving camera datasets difficult to achieve (Tab. \ref{tab:tapvid3D_results}). We have generated moving camera versions of Multi-View Kubric and Multi-View Point Odyssey and plan to add moving camera support to Tracktail. Expanding our training set to include these datasets will hopefully improve robustness and performance on moving camera datasets from TAPVid-3D (Appendix \cref{sec:tapvid3d}).

While Tracktail demonstrates a higher ability for long-range tracking on Multi-View Point Odyssey compared to baseline methods, there is still room for improvement. Highly accurate tracking is necessary for most applications, so reducing point drift is important. Memory-based encoders and cross-attention modules have proven useful in the segmentation space, so they could be useful for multi-view point trackers too \cite{ravi2025sam2,carion2025sam3,jung2026tapnextwhatstrackingpoint}. We hope that releasing the Multi-View Point Odyssey benchmark enables more development to improve long-range point tracking. 

Many point tracking models are trained on purely synthetic data. While this is great for generating large datasets with accurate ground truth, there is still a gap between synthetic and real-world data distributions that limits the performance of these point tracking models on real-world data. We are currently using our pretrained model as a foundation for finetuning on a diverse collection of real-world animal datasets. We are planning for this model to be part of a framework that will reduce the amount of manual keypoint annotation that is needed to get accurate pose tracking and advance the field of multi-view animal tracking and pose estimation. Even without finetuning on real-world animal datasets, Tracktail shows promising results on Rat7M (Fig.~\ref{fig:rat7m}) that make us hopeful for the future of this field.

\section{Conclusion}
\label{sec:conclusion}

Tracktail is a multi-view point tracker that accurately tracks 3D query points without depth input. Our method advances the state of the art by enabling tracking on datasets that don't have ground truth depth without relying on inaccurate depth estimates. Additionally, it addresses the scene normalization problem with a per-camera scene scale computed from the camera geometry, allowing for robust predictions regardless of dataset or scene scale. We also introduce the Multi-View Point Odyssey benchmark dataset, which consists of naturalistic human and animal data for measuring the long-range tracking performance of 3D point trackers. We hope that our model and this benchmark will advance the field for long-term tracking developments and push tracking performance on real-world data even further.

\section*{LLM Usage Statement}

We used Anthropic's Claude models through the Claude Code and Pi coding harnesses throughout this project.
For the model and training code, the authors wrote the core implementation, and the agents assisted with parts of the architecture, losses, and data pipeline, with refactoring, and with debugging.
For research design, LLM agents explored design variants and ran small toy experiments, for example for the output heads and scene normalization. All ideas were generated or reviewed by the authors, who made all design decisions.
For writing, LLM agents assisted with drafting parts of the methods, experiments, and appendix.
The authors reviewed and edited all text, reviewed and tested all code, and checked all citations against their sources. The authors take full responsibility for the content of this paper.

{
    \small
    \bibliographystyle{ieeenat_fullname}
    \bibliography{main}
}

\clearpage
\setcounter{page}{1}
\twocolumn[
    \centering
    \Large
    \textbf{\thetitle}\\
    \vspace{0.5em}Appendix \\
    \vspace{1.0em}
]
\appendix

\section{Multi-View Point Odyssey details}
\label{sec:supp_po}

\paragraph{Rendering and assets} We extend the Point Odyssey generation pipeline \cite{zheng2023point} and follow their protocol to collect data assets for the subjects, environments, scenes, and objects. Scenes are rendered in Blender 3.3 from 9 static cameras placed around the subject, at $960 \times 540$ pixels and 30 fps. For every surface point, we export its ground-truth 3D trajectory and its visibility in each camera. The pipeline produces ground truth depth and segmentation masks as well.

For the human dataset, we downloaded a 108 humanoid characters from Mixamo \cite{mixamo} and 11 of the 24 AMASS datasets \cite{mahmood2019amass}. The generation pipeline randomly selects one of these characters and animates them with SMPL-X \cite{pavlakos2019smplx} motions from a random AMASS sequence.  For the animal dataset, we downloaded DeformingThings4D, which contains 26 animal models and 1,772 animation sequences \cite{li20214dcomplete}. For each scene, the generation pipeline randomly samples an animal model, applies one of 17 fur textures from BlenderKit \cite{blenderkit}, and concatenates 15 randomly sampled animation clips into one long motion. 

For the scenes, we downloaded a collection of 220 4k indoor and outdoor HDRI environment maps from PolyHaven \cite{polyhaven}. The generation pipeline randomly selects an environment as well as 5 distractor objects from Google Scanned Objects \cite{downs2022gso} and PartNet \cite{mo2019partnet}. Throughout each video, the objects are pushed by random forces to ensure movement.

\paragraph{Dataset splits} The human dataset has 113 training sequences (249--2,034 frames) and 8 test sequences. The animal dataset has 210 training and 10 test sequences. Each animal training sequence is a 256-frame window taken from the full rendered sequence of 589--2,815 frames based on the span of 256 frames with the most 3D motion. All test sequences are 499 frames long.

\paragraph{Track cleaning} We found that some generated tracks move extremely fast and out of the scene, so we removed them from the dataset. A track is removed if, in any camera, its projection moves more than 100 pixels between consecutive frames or its depth falls below 1 scene unit. It is also removed if it moves farther from the scene center than 20 times the median distance of points to the center in the first frames. As a result, we discarded a median of 0.2\% of tracks per human sequence and 75\% per animal sequence.

\paragraph{Evaluation tracks} We select 512 tracks per test sequence, stratified by speed. Eligible tracks are visible in at least one of the 4 evaluation cameras (views 2, 4, 5, and 7), in both frames of at least 50 consecutive-frame pairs. We rank them by their mean 3D speed over visible frames. We then sample 128 slow tracks (below the 25th percentile), 256 medium tracks (25th to 75th percentile), and 128 fast tracks (75th to 95th percentile). We excluded the fastest 5\% of tracks because they are much faster than the majority and likely dominated by tracking artifacts. The tracks are used for the 4-camera and 9-camera evaluations.

\section{Architecture details}
\label{sec:supp_architecture}

\paragraph{Video encoder.} Each view is processed independently by the V-JEPA 2.1 ViT-L/16 encoder as a clip of 24 frames at $256 \times 256$ pixels, giving $12 \times 16 \times 16$ spatio-temporal tokens per view (tubelet size 2, patch size 16). We use hierarchical features: the outputs of blocks 6, 12, 18, and 24 are concatenated and linearly projected to 512 dimensions. We add a learned spatial positional embedding to the projected tokens. Temporal position is provided by rotary embeddings in the decoder cross-attention.

\paragraph{Query encoder.} The query encoder computes an embedding independently for each camera, query point, and target time. For camera $c$, the query point $\mathbf{q}_n$ is projected into the image, and we embed:
(i) a $9 \times 9$ pixel patch around the projection in frame $t_n$, encoded with a small CNN,
(ii) the normalized 2D projected position,
(iii) the log of the query's distance to the camera divided by $s_c$,
(iv) the camera's principal point and focal length, normalized by the image size,
(v) the query time $t_n$, the target time $\tau$, and their difference $\tau - t_n$,
(vi) whether the projection lies inside the image, and
(vii) a per-camera occlusion state at the query frame (visible, occluded, or unknown).
Continuous quantities are embedded with Fourier features followed by a linear layer, and discrete ones with learned embeddings. The resulting terms are combined by a learned gated sum, in which per-term sigmoid weights are predicted from all terms, followed by an MLP that produces the 512-dimensional embedding $\mathbf{z}_{n,\tau}^c$.

\paragraph{Occlusion state.} The occlusion state lets the model use visibility information when it is available. In all reported experiments it is set to unknown at the query frame, and it is only filled in from the model's own predictions when tracking across windows. During training it is randomly set to unknown with probability 0.5.

\paragraph{2D queries.} The same network also accepts 2D pixel queries in a single view, which we use as an auxiliary training task. For 2D queries the depth term is replaced by a learned token. The decoder is conditioned on the query type (2D or 3D), and each query type has its own output heads.

\paragraph{Decoder.} The decoder consists of 6 blocks with width 512 and 8 attention heads. Each block applies camera self-attention, temporal self-attention, query-scene cross-attention, and a feed-forward MLP (expansion ratio 4).
For camera self-attention, the embeddings $\{\mathbf{z}_{n,\tau}^c\}_{c=1}^{C}$ of each query point and target time across all cameras attend to each other, with the ray encodings $\mathbf{r}^c_n$ as relative positional encodings as in PRoPE \cite{li2025camerasrelativepositionalencoding}.
For temporal self-attention, the embeddings of all target times of each camera and query point attend to each other, with 1D rotary position embeddings over the frame index.
For query-scene cross-attention, the query embeddings of camera $c$ attend to all tokens of its scene representation $\mathbf{F}^c$. We apply 1D temporal rotary embeddings to the queries (by target time) and keys (by token frame), so that attention is biased by the relative time between the target frame and each video token.

\paragraph{Output heads.} After the last block, the per-camera embeddings are passed through separate linear heads:
\begin{itemize}
    \item \textbf{3D}: the motion of the point relative to its query position, expressed in a ray-local frame aligned with the camera's central ray and normalized by $s_c$ times the image width, with 1024 logarithmically spaced bins per axis.
    \item \textbf{Depth}: the log-depth along the ray, relative to $s_c f_c$.
    \item \textbf{2D}: the absolute pixel position in the view (256 bins per axis).
    \item \textbf{Visibility}: a per-camera visibility logit.
    \item \textbf{3D fusion weight}: a per-camera logit.
\end{itemize}
The 3D motion is mapped back to world coordinates and added to the query position to obtain the per-camera 3D prediction $\mathbf{p}_{n,\tau}^c$. The fusion weights are not supervised directly and are learned end-to-end through the 3D losses.

\paragraph{Bin decoding.} All positional heads are per-axis classifiers, decoded by a soft-argmax restricted to a window of 60 bins around the argmax, with a learned temperature.
The 3D head predicts the motion of the point relative to its query position in a ray-local frame of the camera's central ray. The motion is divided by $s_c W$, where $W = 256$ is the input width, and binned with 1024 bins per axis over $[-1.8, 1.8]$. The bins are uniform in the signed-log space $\operatorname{sign}(x)\log(1+|x|/0.1)$, and the soft-argmax is computed in that space.
The depth head predicts $\log(d / (s_c f_c))$ with 1024 uniform bins over $[-2.5, 2.5]$, where $d$ is the distance along the ray and $f_c$ the focal length of the resized crop.
The 2D head predicts the absolute pixel position in the $256 \times 256$ input with one bin per pixel.
Besides the fused direct 3D prediction, two auxiliary 3D estimates are formed for supervision only. The first back-projects the per-camera 2D predictions along their rays to the predicted depths and averages them with per-camera 2D weights. The second triangulates the per-camera 2D predictions with the same weights.

\section{Evaluation metrics}
\label{sec:supp_eval}

We follow the evaluation protocol of MVTracker \cite{rajic2025mvtracker} and compute all metrics in world coordinates.
A point is considered visible at a frame if it is visible in at least one of the input cameras. Each track is evaluated only on the frames after its query frame.

\paragraph{Position accuracy ($\delta_\text{avg}$).} $\delta_\text{avg}$ is the fraction of visible frames whose 3D error is below a threshold, averaged over a set of thresholds:
$\{1,2,5,10,20\}$\,cm for DexYCB, $\{5,10,20,40\}$\,cm for Panoptic Studio, $\{0.05,0.1,0.2,0.4,0.8\}$ scene units for Multi-View Kubric, and $\{0.3,0.6,1.2,2.4,4.8\}$ scene units for Multi-View Point Odyssey.

\paragraph{Occlusion accuracy (OA).} OA is the fraction of frames in which the predicted visibility matches the ground truth.

\paragraph{Average Jaccard (AJ).} AJ jointly measures position and visibility. At each threshold it is $\text{TP}/(\text{TP}+\text{FP}+\text{FN})$, where a true positive is a frame that is predicted visible, visible in the ground truth, and within the threshold. AJ is averaged over the same thresholds as $\delta_\text{avg}$.

\paragraph{Median trajectory error (MTE).} MTE is the median 3D distance between the predicted and ground-truth positions over the visible frames of a track. The units used for MTE on each dataset are described at the end of this section.

\paragraph{Survival rate.} For the long Point Odyssey sequences we also report the survival rate: the fraction of tracks whose 3D error has not yet exceeded a threshold at any visible frame up to time $t$.

\paragraph{Aggregation.} Metrics are computed per track, averaged over the tracks of each sequence, and then averaged over sequences. Tracks with fewer than two visible frames from the query frame on are skipped. MTE uses the lower median for an even number of frames. On Multi-View Kubric and DexYCB, we report the mean of the static and dynamic track scores, following MVTracker: a track is static if its ground-truth path length is below 0.01 and dynamic if it is above 0.1 (Kubric units and meters, respectively), and tracks in between count in neither group.
Points for which a method produces no prediction (\eg failed triangulation) count as predicted occluded for AJ, $\delta_\text{avg}$, and OA, and are skipped for MTE and survival.

\paragraph{Visibility thresholds.} Each method uses its own threshold on the predicted visibility probability. For Tracktail we use 0.1. For MVTracker we use its official thresholds (0.5 on Multi-View Kubric and 0.01 on DexYCB and Panoptic Studio) and 0.5 on Multi-View Point Odyssey. For ARROW we use 0.5. The CoTracker3 baseline uses the binary visibility returned by CoTracker3, which applies its default threshold of 0.9. The visibility threshold only affects AJ and OA.

\paragraph{Units.} All errors are computed in the world coordinates of each dataset. Following MVTracker \cite{rajic2025mvtracker}, MTE is reported in centimeters for Panoptic Studio and DexYCB, whose coordinates are in meters. Multi-View Kubric coordinates are not metric. We multiply its errors by 13, which reproduces the scale of the Kubric MTE values reported by MVTracker. 

The $\delta_\text{avg}$ thresholds for Multi-View Kubric ($\{0.05,0.1,0.2,0.4,0.8\}$) are in unscaled Kubric units. Multi-View Point Odyssey errors, $\delta_\text{avg}$ thresholds, and survival thresholds are in Point Odyssey scene units.

\section{Training}
\label{sec:supp_training}

\subsection{Sampling and augmentation.} Each training sample is a 24-frame clip from 1--8 randomly chosen cameras with 128 query points. 95\% of the query points are moving points (faster than 3 pixels per frame).
Each view is cropped to the bounding box of the projected points and resized to 256 pixels on its longer side, as at inference.
Query times are sampled from the frames in which the point is visible, strongly favoring the first and last frame of the clip, so the model learns to track both forward and backward in time.
With probability 0.25 we instead use a single camera with 2D pixel queries.
We apply the following augmentations:
\begin{itemize}
    \item a random global 3D rotation of the scene and cameras.
    \item with probability 0.4, restricting the query points to a random 3D sphere, whose radius is 5--75\% of the scene extent, to produce zoomed-in views.
    \item per-camera in-plane rotations ($\pm 45^\circ$), cutout occlusions, blur, color, gamma, and noise augmentations, and grayscale conversion.
    \item setting the occlusion state inputs to unknown with probability 0.5.
\end{itemize}

\subsection{Losses}
\label{sec:supp_losses}

The training loss is a weighted sum of 3D, auxiliary, classification, visibility, and smoothness terms:
\begin{align}
\mathcal{L} ={}& \lambda_\text{3D}\mathcal{L}_\text{3D} + \lambda_\text{cam}\mathcal{L}_\text{cam} + \lambda_\text{occ}\mathcal{L}_\text{occ} \nonumber \\
&+ \lambda_\text{2D}\mathcal{L}_\text{2D} + \lambda_\text{d}\mathcal{L}_\text{depth} + \textstyle\sum_{k} \lambda_\text{rep}^{k}\mathcal{L}_\text{rep}^{k} \nonumber \\
&+ \textstyle\sum_{m} \lambda_\text{CE}^{m}\mathcal{L}_\text{CE}^{m} + \lambda_\text{vc}\mathcal{L}_\text{vis}^\text{cam} + \lambda_\text{v}\mathcal{L}_\text{vis} \nonumber \\
&+ \lambda_\text{s}\left(\mathcal{L}_\text{sm}^\text{3D} + \mathcal{L}_\text{sm}^\text{2D}\right),
\end{align}
with $k \in \{\text{fused}, \text{rays}, \text{tri}\}$ and $m \in \{\text{3D}, \text{2D}, \text{depth}\}$. The weights are listed in \cref{tab:loss_weights}.

Let $\rho_\delta$ be the Huber loss applied per coordinate, $\hat{\mathbf{p}}_{n,\tau}$ and $\hat{\mathbf{p}}^c_{n,\tau}$ the fused and per-camera 3D predictions, $\mathbf{p}_{n,\tau}$ the ground truth, $v_{n,\tau}$ and $v^c_{n,\tau}$ the overall and per-camera ground-truth visibility, and $\pi_c$ the projection into camera $c$. Let $\bar{s}$ be the median of $s_c$ over cameras. All sums are normalized by the number of terms, and constant normalization factors are omitted.

\paragraph{3D losses.} The fused 3D prediction is supervised on visible and occluded frames separately, and the per-camera predictions on the frames visible in that camera. With the scale-normalized errors $\mathbf{e}_{n,\tau} = (\hat{\mathbf{p}}_{n,\tau} - \mathbf{p}_{n,\tau})/\bar{s}$ and $\mathbf{e}^c_{n,\tau} = (\hat{\mathbf{p}}^c_{n,\tau} - \mathbf{p}_{n,\tau})/s_c$,
\begin{align}
\mathcal{L}_\text{3D} &= \textstyle\sum_{n,\tau} v_{n,\tau}\, \rho_\delta(\mathbf{e}_{n,\tau}), \\
\mathcal{L}_\text{occ} &= \textstyle\sum_{n,\tau} (1 - v_{n,\tau})\, \rho_\delta(\mathbf{e}_{n,\tau}), \\
\mathcal{L}_\text{cam} &= \textstyle\sum_{c,n,\tau} v^c_{n,\tau}\, \rho_\delta(\mathbf{e}^c_{n,\tau}).
\end{align}

\paragraph{Auxiliary losses.} Let $\hat{\mathbf{u}}^c$ and $\hat{d}^c$ be the per-camera 2D and depth predictions, and $\mathbf{u}^c = \pi_c(\mathbf{p})$ and $d^c$ the ground truth. The 2D, depth, and reprojection losses are computed on the frames visible in each camera:
\begin{align}
\mathcal{L}_\text{2D} &= \textstyle\sum_{c,n,\tau} v^c_{n,\tau}\, \rho\big(\hat{\mathbf{u}}^c_{n,\tau} - \mathbf{u}^c_{n,\tau}\big), \\
\mathcal{L}_\text{depth} &= \textstyle\sum_{c,n,\tau} v^c_{n,\tau}\, \rho_\delta\Big(\frac{\hat{d}^c_{n,\tau} - d^c_{n,\tau}}{s_c f_c}\Big), \\
\mathcal{L}_\text{rep}^{k} &= \textstyle\sum_{c,n,\tau} v^c_{n,\tau}\, \rho_{16}\big(\pi_c(\hat{\mathbf{p}}^{k}_{n,\tau}) - \mathbf{u}^c_{n,\tau}\big).
\end{align}
Here $\hat{\mathbf{p}}^\text{fused}$ is the fused 3D prediction, $\hat{\mathbf{p}}^\text{rays}$ is the average of the per-camera 2D predictions back-projected to their predicted depths, and $\hat{\mathbf{p}}^\text{tri}$ is the weighted triangulation of the per-camera 2D predictions.

\paragraph{Classification losses.} $\mathcal{L}_\text{CE}^{m}$ is the cross-entropy between the bin logits of head $m$ and the bin of the ground truth, on visible frames.

\paragraph{Visibility losses.} With the binary cross-entropy
\begin{equation}
\text{BCE}(\hat{v}, v) = -v \log \hat{v} - (1 - v) \log(1 - \hat{v}),
\end{equation}
where $\hat{v}$ is a predicted visibility probability and $v \in \{0, 1\}$ the ground truth,
\begin{align}
\mathcal{L}_\text{vis}^\text{cam} &= \textstyle\sum_{c,n,\tau} \text{BCE}(\hat{v}^c_{n,\tau}, v^c_{n,\tau}), \\
\mathcal{L}_\text{vis} &= \textstyle\sum_{n,\tau} \text{BCE}(\hat{v}_{n,\tau}, v_{n,\tau}),
\end{align}
where $\hat{v}^c_{n,\tau}$ are the per-camera and $\hat{v}_{n,\tau}$ the noisy-OR visibility predictions.

\paragraph{Smoothness losses.} Let $\Delta^4$ be the 4th-order temporal difference, computed per coordinate on windows of visible frames. The smoothness loss penalizes the part of the predicted $\Delta^4$ that exceeds 1.5 times that of the ground truth:
\begin{equation}
\mathcal{L}_\text{sm}^\text{3D} = \textstyle\sum_{n,\tau} \frac{1}{\bar{s}} \big[\, |\Delta^4 \hat{\mathbf{p}}_{n,\tau}| - 1.5\,|\Delta^4 \mathbf{p}_{n,\tau}| \,\big]_+,
\end{equation}
where $[x]_+ = \max(0, x)$. $\mathcal{L}_\text{sm}^\text{2D}$ is defined analogously on the per-camera 2D predictions in pixels.

\paragraph{2D queries.} For 2D queries, where the scene scale is undefined, the 3D positions and depths in these losses are instead expressed relative to the camera center and divided by the mean depth of the visible ground-truth points, so the 3D predictions are supervised up to scale, and the reprojection losses are omitted.

\begin{table}[h]
\centering
\caption{\textbf{Loss weights.} The Huber threshold $\delta$ of the 3D and depth losses is 6 during finetuning and 4 during pretraining.}
\label{tab:loss_weights}
\begin{tabular}{lc lc}
\toprule
Weight & Value & Weight & Value \\
\midrule
$\lambda_\text{3D}$ & 0.2 & $\lambda_\text{rep}^\text{tri}$ & 0.0625 \\
$\lambda_\text{cam}$ & 0.1 & $\lambda_\text{CE}^\text{3D}$ & 0.4 \\
$\lambda_\text{occ}$ & 0.1 & $\lambda_\text{CE}^\text{2D}$ & 0.6 \\
$\lambda_\text{2D}$ & 1.5 & $\lambda_\text{CE}^\text{depth}$ & 1.5 \\
$\lambda_\text{d}$ & 0.5 & $\lambda_\text{vc}$ & 5.0 \\
$\lambda_\text{rep}^\text{fused}$ & 0.0625 & $\lambda_\text{v}$ & 1.0 \\
$\lambda_\text{rep}^\text{rays}$ & 0.003125 & $\lambda_\text{s}$ & 0.5 \\
\bottomrule
\end{tabular}
\end{table}

\begin{table*}[t]
\centering
\caption{\textbf{Analysis of CoTracker3 Baseline triangulation failures} - For each dataset, we calculated the percentage of (track, frame) pairs for which the CoTracker3 baseline could not produce a prediction due to failed triangulation. We break down this percentage into two categories based on the cause of the failure. \textit{Not visible enough} means that the point was not visible in enough camera views required to triangulate ($<2$). This includes frames where the point is occluded in all views, for which no prediction is the correct output. \textit{Visible enough} means that the point was visible in enough camera views to triangulate ($\geq$ 2), but CoTracker3 didn't predict that the point was visible in enough views to triangulate. The substantial percentage of failed points in each dataset demonstrates that this naive baseline is not sufficient for most applications.}
\label{tab:cotracker_triangulation}
\begin{tabular}{llccc}
\toprule
 & & & \multicolumn{2}{c}{Cause of the triangulation failures (\%)} \\
\cmidrule(lr){4-5}
Dataset & Number of Views & Total failures (\%) & Not visible enough & Visible enough \\
\midrule
\rowcolor{rowgray}
Multi-View Kubric       & 4 & 40.9 & 60.5 & 39.5 \\
DexYCB                 & 4 & 10.6 & 64.4 & 35.6 \\
\rowcolor{rowgray}
Panoptic Studio        & 4 & 20.2 & 51.4 & 48.6 \\
\midrule
Multi-View Point Odyssey - Human   & 4 & 48.7 & 26.7 & 73.3 \\
\rowcolor{rowgray}
Multi-View Point Odyssey - Human   & 9 & 29.6 &  7.3 & 92.7 \\
Multi-View Point Odyssey - Animal  & 4 & 52.0 & 28.8 & 71.2 \\
\rowcolor{rowgray}
Multi-View Point Odyssey - Animal  & 9 & 30.4 &  4.6 & 95.4 \\
\bottomrule
\end{tabular}
\end{table*}

\section{Additional ablations}
\label{sec:supp_ablation}

Supplementary Tables \ref{tab:ablation_encoder} and \ref{tab:ablation_latent} report the video encoder size and latent dimension ablations discussed in the main text.

\paragraph{Effect of the number of input camera views.}
For each number of views $N$, we use a separate set of 512 tracks, randomly sampled from the tracks that are visible in at least 2 frames in any of the $N$ views, with a quarter of them queried at a random visible frame and the rest at their first visible frame, as in MVTracker, so the 4-view values differ from \cref{tab:main_results} in main text.

\begin{figure}[t]
    \centering
    \includegraphics[width=\linewidth]{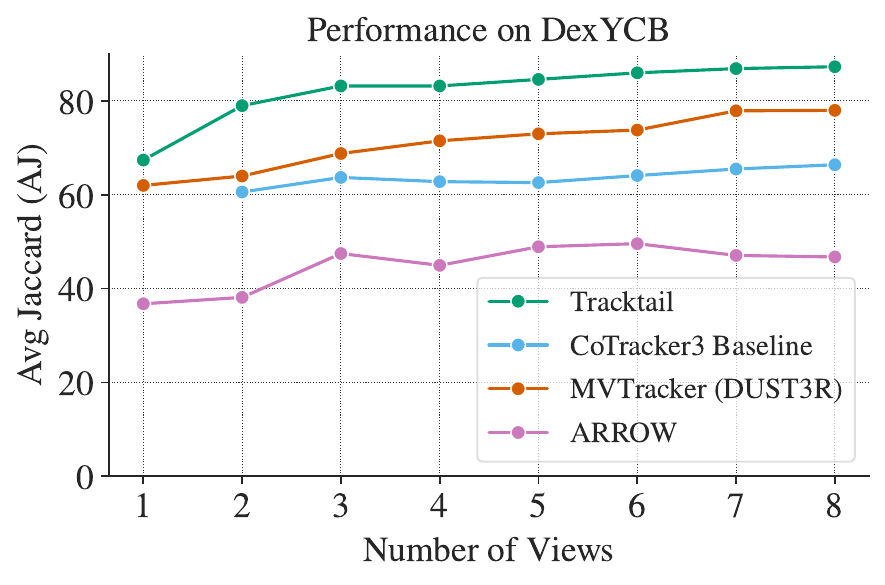}
    \caption{\textbf{Effect of the number of input camera views} - Average Jaccard (AJ) of Tracktail and baseline methods on DexYCB given 1--8 views. AJ of Tracktail and MVTracker generally improves with more views, reaching 87.3 and 78.0, respectively. ARROW and the CoTracker3 baseline gain little from more views. }
    \label{fig:ncams}
\end{figure}

\begin{table}[t]
\centering
\caption{\textbf{Varying video encoder size} - Tracktail models with different V-JEPA 2.1 video encoder sizes, evaluated on Multi-View Kubric. Increasing the encoder size improves all metrics. The largest gain comes from going from the base (86M parameters) to the large (300M) encoder, and the giant encoder (1B) gives a further improvement. The gigantic encoder (2B) gives only a marginal further gain in AJ and $\delta_\text{avg}$, with the same OA and MTE as the giant encoder. All models were trained with the pretraining stage only (800k iterations, frozen encoder). The large model is our pretrained model before finetuning.}
\label{tab:ablation_encoder}
\begin{tabular}{l cccc}
\toprule
\multirow{2}{*}{Encoder Size}
  & \multicolumn{4}{c}{Multi-View Kubric~\cite{rajic2025mvtracker}} \\
\cmidrule(lr){2-5}
  & AJ$\uparrow$ & $\delta_\text{avg}\uparrow$ & OA$\uparrow$ & MTE$\downarrow$\\
\midrule
\rowcolor{rowgray}
Base (86M) & 56.5 & 66.6 & 88.7 & 3.4 \\
Large (300M) & 60.2 & 70.4 & 89.4 & 2.8 \\
\rowcolor{rowgray}
Giant (1B) & 61.7 & 71.8 & \textbf{89.5} & \textbf{2.6} \\
Gigantic (2B) & \textbf{62.1} & \textbf{72.4} & \textbf{89.5} & \textbf{2.6} \\
\bottomrule
\end{tabular}
\end{table}

\begin{table}[t]
\centering
\caption{\textbf{Varying latent dimension} - Tracktail models with decoder latent dimensions of 256, 512, and 1024, evaluated on Multi-View Kubric. A latent dimension of 512 gives the best performance on all metrics, with the same occlusion accuracy as 1024. All models were trained with the pretraining stage only (800k iterations, frozen encoder).}
\label{tab:ablation_latent}
\begin{tabular}{c cccc}
\toprule
\multirow{2}{*}{Latent Dim }
  & \multicolumn{4}{c}{Multi-View Kubric~\cite{rajic2025mvtracker}} \\
\cmidrule(lr){2-5}
  & AJ$\uparrow$ & $\delta_\text{avg}\uparrow$ & OA$\uparrow$ & MTE$\downarrow$ \\
\midrule
\rowcolor{rowgray}
256 & 56.2 & 66.4 & 88.6 & 3.2 \\
512 & \textbf{60.2} & \textbf{70.4} & \textbf{89.4} & \textbf{2.8} \\
\rowcolor{rowgray}
1024 & 59.9 & 70.2 & \textbf{89.4} & 2.9 \\
\bottomrule
\end{tabular}
\end{table}

\section{Survival rate}
\label{sec:supp_survival}

\begin{figure*}[t]
    \centering
    \includegraphics[width=\linewidth]{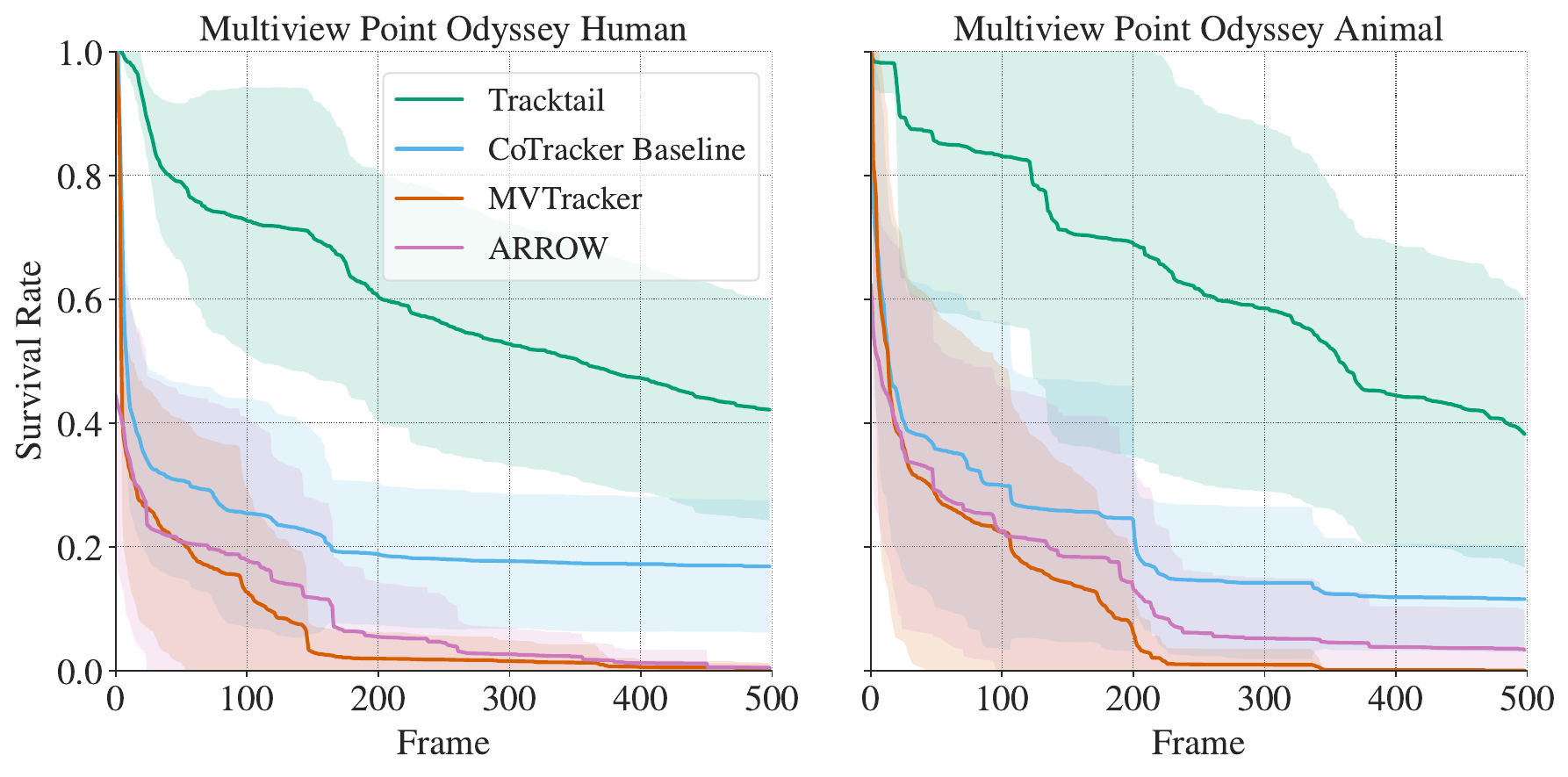}
    \caption{\textbf{Survival rate for long-term tracking} - Survival rate of points over the 499 frames of the Multi-View Point Odyssey Human (left) and Animal (right) test sets with 4 cameras. A track fails at the first visible frame where its 3D error exceeds 1.2 scene units, and the survival rate is the fraction of tracks that have not failed. The shaded regions show the standard deviation across test sequences. Tracktail's survival rate decreases gradually over time, while the CoTracker3 baseline, ARROW, and MVTracker (DUSt3R depth) drop sharply within the first 25 frames. See \cref{fig:po_survival_tracktail} for the effect of the threshold on Tracktail.}
    \label{fig:po_survival_comparison}
\end{figure*}

Figure \ref{fig:po_survival_comparison} shows the survival of points on Multi-View Point Odyssey across all 499 frames for Tracktail and the baseline methods. Tracktail maintains a high survival rate for the first 50 frames, while the other methods exhibit a sharp decrease in survival almost immediately. For the entire 499-frame duration, Tracktail's survival rate is far above the other methods for the, decaying to just over 40\% survival by the final frame. Both MVTracker and ARROW decay to 0\% survival, while the CoTracker3 baseline plateaus at around 18\% survival.  

\begin{figure*}[t]
    \centering
    \includegraphics[width=\linewidth]{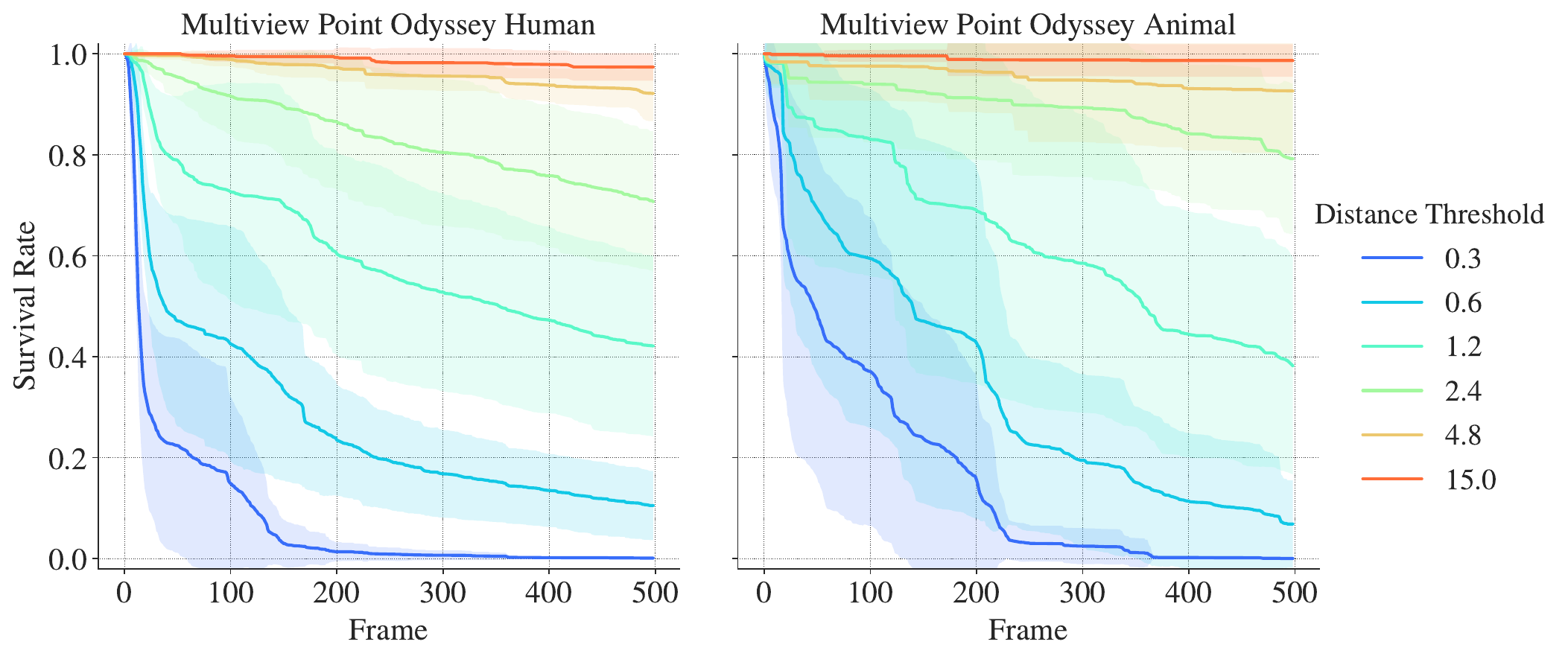}
    \caption{\textbf{Effect of distance threshold on survival rate} - We calculated the survival rate of Tracktail on the Multi-View Point Odyssey Human (left) and Animal (right) test datasets with 4 cameras, for various distance thresholds in Point Odyssey scene units. The error regions show the standard deviation across the test trials. Lower distance thresholds result in a larger decrease of survival rate across the 499 frames. We chose a distance threshold of 1.2 for Fig. \ref{fig:po_survival_comparison} in the main paper because it shows a good balance of tracking going from good to poor over the 499 frames.}
    \label{fig:po_survival_tracktail}
\end{figure*}

We tested various 3D distance thresholds for the survival rate and picked a threshold of 1.2 scene units, as it shows Tracktail decaying from good to poor performance. Figure \ref{fig:po_survival_tracktail} shows the effect of different thresholds on the survival rate for Tracktail. More relaxed thresholds allows tracking to stay between 80-100\% survival, while more stringent thresholds show performance decaying to 0-20\% survival for the majority of frames.

\section{Evaluation on TAPVid-3D}
\label{sec:tapvid3d}

In Table \ref{tab:tapvid3D_results}, we report camera-coordinate 3D tracking performance of Tracktail, ARROW \cite{fradlin2026arrow}, and D4RT \cite{zhang2026d4rt} on the TAPVid-3D datasets \cite{koppula2024tapvid3d}. While D4RT achieves the highest performance on Panoptic Studio, Tracktail outperforms ARROW for both the AJ and APD$_\text{3D}$ metrics. However, both D4RT and ARROW achieve much higher performance than Tracktail on DriveTrack and Aria Digital Twin. Given Figure \ref{fig:ncams}, we found that Tracktail exhibits the largest jump in performance when going from one to two input views. Thus, the monocular setting of the TAPVid-3D datasets makes them challenging. Additionally, DriveTrack and Aria Digital Twin are both moving camera datasets. D4RT and ARROW included moving camera datasets in their training mixtures, while Tracktail was only trained on static camera sequences. This likely explains how D4RT and ARROW were able to achieve much higher performance on these datasets.

   \begin{table*}[t!]
   \centering
   \caption{\textbf{Comparison of 3D point tracking methods on TAPVid-3D~\cite{koppula2024tapvid3d}} - Camera-coordinate 3D tracking on the three real-world
 monocular datasets of TAPVid-3D (mini-val split, full sequences, official evaluation with per-sequence median scale alignment). Tracktail is given the ground-truth 3D query points and intrinsics, treats each video as a static camera, tracks forward and backward from the query frame, and uses a visibility threshold of 0.1. The occlusion state input at the query frame is unknown. The ADT results are computed on 49 of the 50 mini-val videos.  GT
 intr.\ indicates whether the method is given the ground-truth camera intrinsics. D4RT and ARROW metrics are taken directly from
 their papers \cite{zhang2026d4rt, fradlin2026arrow}. Tracktail improves upon ARROW on Panoptic Studio, but still lags behind D4RT. Tracktail performs poorly on DriveTrack and ADT as expected, as those datasets have moving cameras.}
   \label{tab:tapvid3D_results}
   \resizebox{\textwidth}{!}{%
   \begin{tabular}{l c ccc ccc ccc}
   \toprule
   \multirow{2}{*}{Method}
     & \multirow{2}{*}{GT intr.}
     & \multicolumn{3}{c}{DriveTrack}
     & \multicolumn{3}{c}{Aria Digital Twin}
     & \multicolumn{3}{c}{Panoptic Studio} \\
   \cmidrule(lr){3-5}\cmidrule(lr){6-8}\cmidrule(lr){9-11}
     &
     & AJ$\uparrow$ & APD$_\text{3D}\uparrow$ & OA$\uparrow$
     & AJ$\uparrow$ & APD$_\text{3D}\uparrow$ & OA$\uparrow$
     & AJ$\uparrow$ & APD$_\text{3D}\uparrow$ & OA$\uparrow$ \\
   \midrule
   D4RT~\cite{zhang2026d4rt}
     & $\times$
     & \underline{0.257} & 0.345 & \textbf{0.875}
     & 0.240 & 0.319 & \textbf{0.926}
     & 0.138 & 0.186 & \textbf{0.897} \\
   ARROW~\cite{fradlin2026arrow}
     & $\times$
     & 0.241 & \underline{0.350} & \underline{0.816}
     & \textbf{0.360} & \textbf{0.473} & \underline{0.902}
     & 0.141 & 0.201 & \underline{0.870} \\
   \midrule
   \rowcolor{rowgray}
   D4RT~\cite{zhang2026d4rt}
     & $\checkmark$
     & \textbf{0.304} & \textbf{0.410} & \textbf{0.875}
     & \underline{0.307} & \underline{0.408} & \textbf{0.926}
     & \textbf{0.372} & \textbf{0.498} & \textbf{0.897} \\
  Tracktail (Ours)
     & $\checkmark$
     & 0.058 & 0.100 & 0.797
     & 0.068 & 0.125 & 0.740
     & \underline{0.228} & \underline{0.353} & 0.797 \\
   \bottomrule
   \end{tabular}%
   }
   \end{table*}

\begin{figure*}[t!]
\centering
\includegraphics[width=\textwidth]{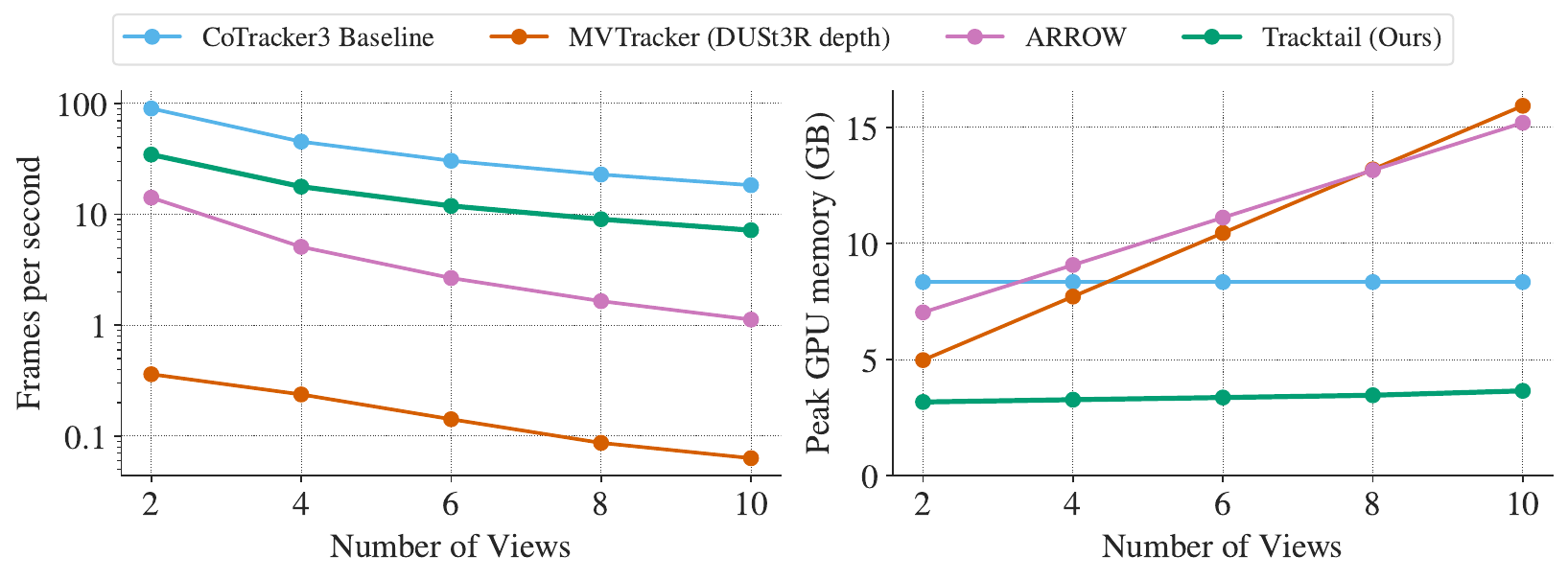}
\caption{\textbf{Inference speed and GPU memory} - Frames per second (left, log scale) and peak GPU memory
allocated (right) as a function of the number of views, tracking 100 query points through a 96-frame  clip on one NVIDIA A100 (80\,GB). Timing covers the full pipeline, from decoded frames in memory to 3D
tracks. Tracktail runs consistently faster and, at 100 query points, uses less GPU memory than ARROW and MVTracker with DUSt3R depth. Each method was run as a separate job on an A100 node. The 10-view setting repeats one of the 9 cameras.}
\label{fig:speed_memory}
\end{figure*}

\begin{figure*}[t!]
\centering
\includegraphics[width=\textwidth]{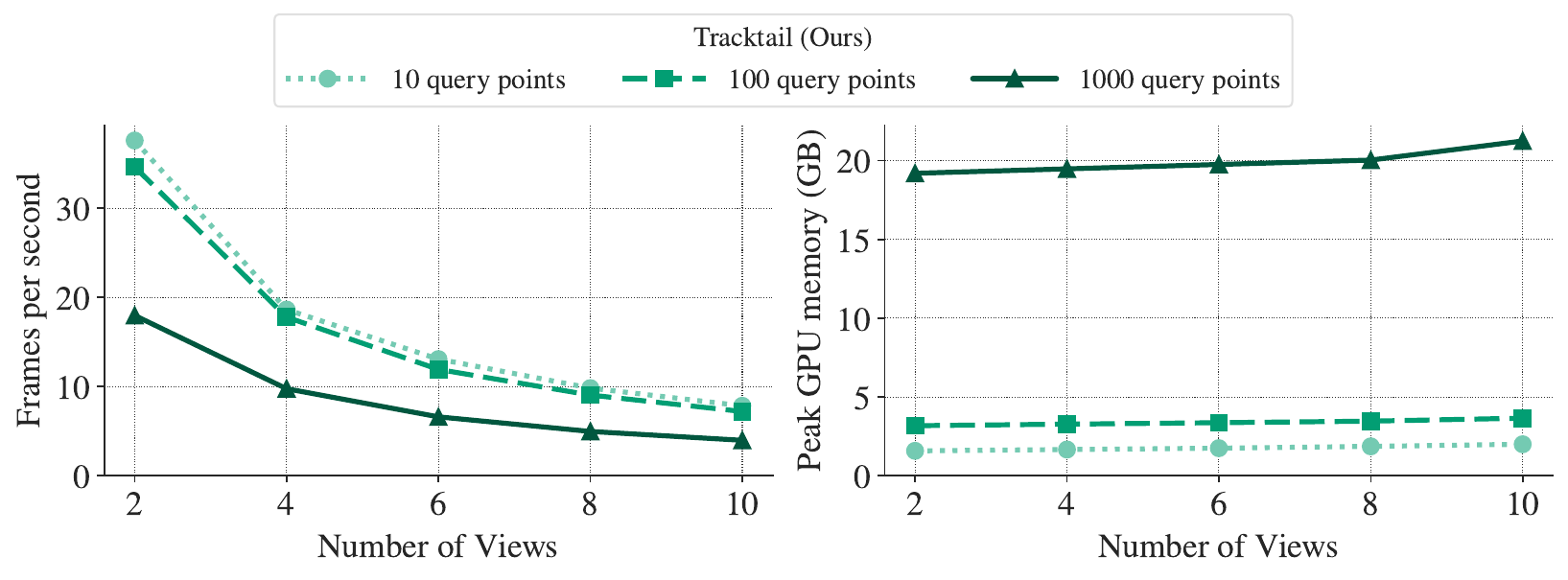}
\caption{\textbf{Tracktail speed and memory with query points} - Frames per second (left) and peak GPU
memory allocated (right) of Tracktail as a function of the number of views, for 10, 100 and 1000 query
points, with the same setup as Fig.~\ref{fig:speed_memory}. The scene encoder processes one view at a time,
so memory grows only slightly with the number of views (e.g.\ 3.2 to 3.7\,GB from 2 to 10 views at 100
points). Memory is driven by the number of query points: about 2\,GB at 10 points and about 20\,GB at 1000
points. Speed is nearly the same for 10 and 100 points and roughly halves at 1000 points.}
\label{fig:speed_memory_tracktail}
\end{figure*}

\section{Cover figure details}
\label{sec:cover_figure}

Fig.~\ref{fig:rat7m} shows Tracktail tracking a dense point cloud on a rat in a 3\,s (360-frame) clip from the Rat7M test set \cite{Dunn2021DANNCE}, recorded by 6 calibrated cameras at 120 fps. Tracktail was not finetuned on Rat7M. Because Tracktail takes 3D query points as input, we first built a point cloud of the rat at the first frame and then tracked every point in it.

\paragraph{Initial point cloud} We segmented the rat in each view with SAM \cite{kirillov2023segment}, prompted by a background-subtraction mask, and intersected the masks to obtain a visual hull. This gives a coarse 3D surface in the units of the calibration. We then predicted per-view depth maps with VGGT \cite{wang2025vggtvisualgeometrygrounded} on crops around the rat. Since these depths are not in metric units, we aligned each map to the visual hull with a robust affine fit and back-projected it using the Rat7M calibration. We kept only points that fell inside the masks of all views and agreed with the depth of at least two other views. We repeated the alignment three times, each time using the fused cloud in place of the hull, and voxel-downsampled the result to 5,668 points.

\paragraph{Tracking and rendering} All points were tracked with Tracktail through the 360 frames from the 6 videos and camera parameters. We discarded 19 static points on the bedding. The right side of Fig.~\ref{fig:rat7m} renders the predicted 3D points at 0, 0.95, and 2.1\,s, with the query frame drawn with larger points. Points are colored from head (red) to tail (pink) along the main axis of the initial cloud, so each color marks the same body location over time. The lines show the trajectories of 200 points chosen by farthest point sampling, lightly smoothed in time for display.

\end{document}